\documentclass{article}
\usepackage{spconf,amsmath,graphicx}
\usepackage{amsmath}
\usepackage{amssymb}
\usepackage{mathtools}
\usepackage{enumerate}
\usepackage{bbm}
\usepackage{algorithm}
\usepackage{algorithmic}
\usepackage{epsfig}
\usepackage{color}
\usepackage{graphicx}
\usepackage{caption}
\usepackage{subcaption}
\usepackage{cases}
\usepackage{url}
\usepackage{cite}
\usepackage{fancyhdr}
\usepackage{tocloft}
\usepackage{pdfpages}
\usepackage{diagbox}
\usepackage{makecell}

\renewcommand{\thefigure}{\arabic{figure}}

\newcounter{lemma}
\newcounter{proposition}
\newcounter{theorem}

\newcommand{\mbc}{\mathbf{c}}

\newcommand{\mbj}{\mathbf{j}}

\newcommand{\mbm}{\mathbf{m}}

\newcommand{\mbs}{\mathbf{s}}
\newcommand{\mbt}{\mathbf{t}}

\newcommand{\mbv}{\mathbf{v}}

\newcommand{\mbx}{\mathbf{x}}

\newcommand{\ignore}[1]{}

\makeatletter
\DeclareRobustCommand\onedot{\futurelet\@let@token\@onedot}
\def\@onedot{\ifx\@let@token.\else.\null\fi\xspace}

\makeatother

\title{HOReeNet: 3D-aware Hand-Object Grasping Reenactment}

\name
 {Changhwa Lee$^{1,2}$, Junuk Cha$^{1}$, Hansol Lee$^{1}$, Seongyeong Lee$^{1,3}$, Donguk Kim$^{1}$, Seungryul Baek$^{1}$}
\address{UNIST, South Korea$^{1}$,         Kakao Brain Corp, South Korea$^{2}$, NC Soft, South Korea$^{3}$}
\begin{document}
\ninept
\maketitle
\begin{abstract}
We present HOReeNet, which tackles the novel task of manipulating images involving hands, objects, and their interactions. Especially, we are interested in transferring objects of source images to target images and manipulating 3D hand postures to tightly grasp the transferred objects. Furthermore, the manipulation needs to be reflected in the 2D image space. In our reenactment scenario involving hand-object interactions, 3D reconstruction becomes essential as 3D contact reasoning between hands and objects is required to achieve a tight grasp. At the same time, to obtain high-quality 2D images from 3D space, well-designed 3D-to-2D projection and image refinement are required. Our HOReeNet is the first fully differentiable framework proposed for such a task. On hand-object interaction datasets, we compared our HOReeNet to the conventional image translation algorithms and reenactment algorithm. We demonstrated that our approach could achieved the state-of-the-art on the proposed task. 
\end{abstract}
%

\section{Introduction}

\begin{figure}[!t]
\centering
\includegraphics[width=1\linewidth]{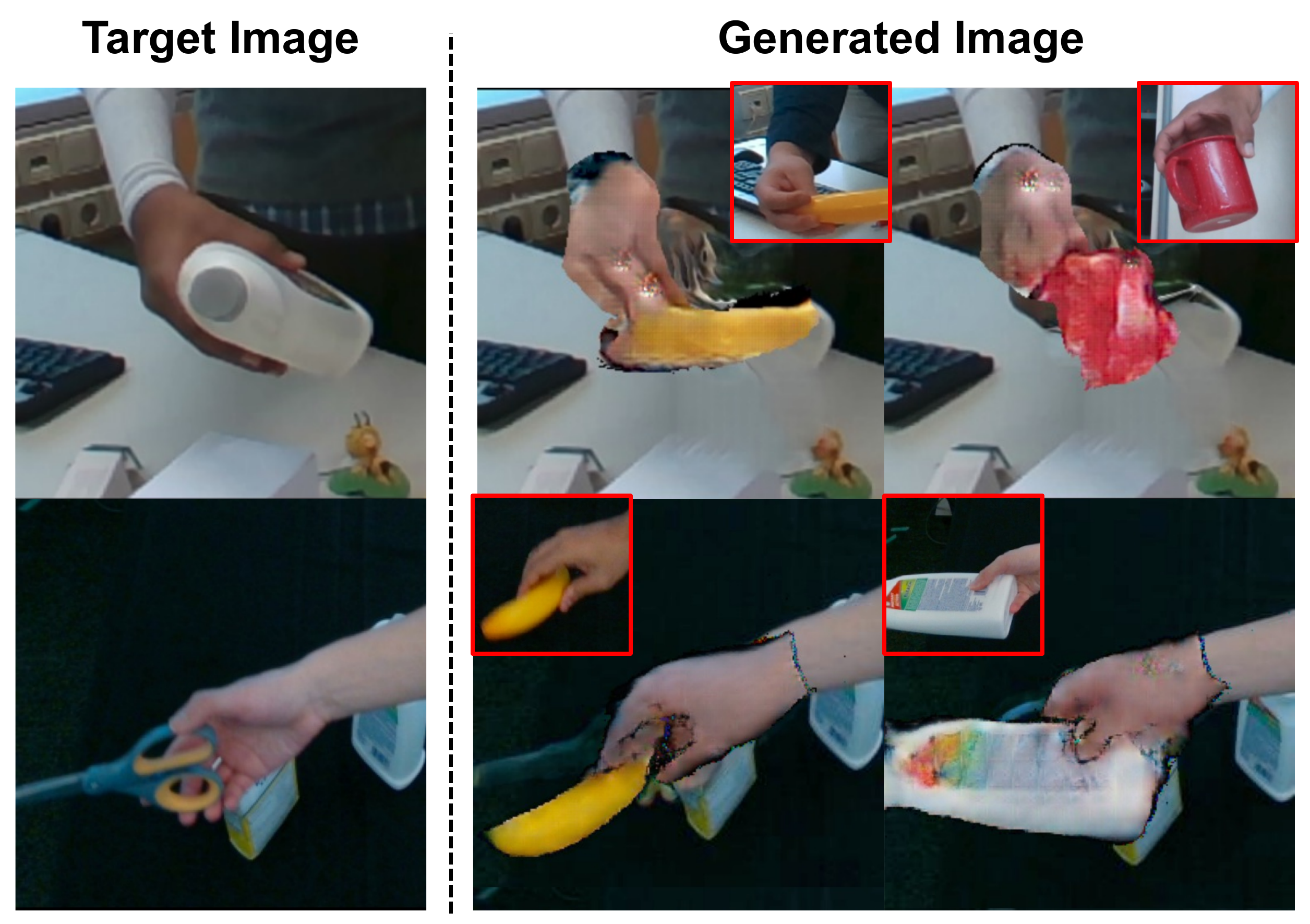}
   \caption{Given a pair of source (red box) and target images, source $\mbx_{\text{src}}$'s object is transferred to the target $\mbx_{\text{tar}}$, hand pose parameters in $\mbx_{\text{tar}}$ are manipulated to tightly grasp the object. Finally, a new image is generated for the new interaction.}
\label{fig:first}
\end{figure}

Simulating and visualizing interacting hands with diverse tools and objects~\cite{interactivevr, vrsensor, Song2004ANP, handtrackdeeplabel} is an important application of virtual reality (VR) and augmented reality (AR). While the simple rendering pipeline of hand and object 3D meshes could achieve the goal; there are few challenges that prevent accomplishing it using commercial 3D rendering pipelines (e.g. Blender, MAYA, ZBrush, etc.): 1) most pipelines work only for objects whose 3D meshes and textures are pre-given; thus types of rendered objects are limited as collecting 3D meshes and textures takes effort, 2) the physical simulator is not involved in the rendering pipelines; while it is essential to simulate the surface contact between hand and object meshes in this scenario, 3) the quality of rendered images frequently have a gap to realistic images. Motivated by the recent success of image manipulation approaches, in this paper, we propose an alternative approach that achieves the scenario of re-enacting hand-object grasps, {\it hand-object grasping reenactment} that transfers objects of an image to the other image, manipulates hand postures to tightly grasp the transferred object 
as in Fig.~\ref{fig:first}. Since we are based on image manipulation, it is possible to involve objects whose 3D meshes and textures are unknown. We also propose the \emph{HOReeNet}, an image synthesis network that learns to re-enact hand-object grasps reflecting the 3D surface contacts and enhancing image qualities.

In our pipeline, we propose to reconstruct both hand and object 3D meshes, manipulate and project them onto the 2D image space via deep learning methods: The entire structure is composed of 4 distinct differentiable modules that 1) estimate 3D meshes of hands and objects from both source and target images, 2) exchange source and target object meshes and manipulate 3D poses of hands to tightly grasp translated objects, 3) project manipulated hand and object meshes into 2D images and refine the quality of projected 2D foreground images. 4) After that, the foreground region of the background image is in-painted and combined with the refined foreground image to generate the final image.


Our contribution in this paper could be summarized as follows:
\begin{itemize}
\setlength\itemsep{-0.2em}
\item We propose to deal with a novel hand-object grasp reenactment task that transfers objects from source to target images, manipulates hand poses of target images for the tight grasp and, generates quality 2D images reflecting 3D manipulation.
\item We present HOReeNet, a fully differentiable framework for tackling such a challenging task. The entire system is engineered to achieve the successful hand-object grasping reenactment effectively.
\item We perform experiments to prove the superiority of our algorithm over other competitive baselines. 
\end{itemize}


\section{Related Works}

\noindent \textbf{Hand pose estimation.}
Hand pose estimation is proposed to map the single images towards 3D skeletal representation~\cite{zimmermann2017learning, baek2018augmented}. Recently, the end-to-end framework for estimating full 3D meshes from RGB images have been established in~\cite{baek2019pushing,baek2020weakly} using the MANO model~\cite{MANO:SIGGRAPHASIA:2017}. Joint reconstruction of hands and objects was dealt with in \cite{yana2019}.


\noindent \textbf{Reenactment.} In facial domain, model-based works~\cite{zhang2020freenet} propose to use facial landmark points to represent the motion (pose and expression). Instead of landmarks, some works~\cite{wu2018reenactgan} tried to use latent features to represent facial identity and motion. \cite{tripathy2020icface} proposed the action unit (AU) that aims at modeling the specific muscle activities and insist that each combination of them can produce different facial expression, which allows editing without identity distortion. In human body domain, clothing reenactment is proposed to transfer a target clothing item onto the corresponding region of a human body~\cite{wang2018toward}. There have been a work to estimate clothing deformations to the target body~\cite{patel2020tailornet}. \cite{knoche2020reposing} proposed a method that infers 3D body structures and manipulates it through geometric warping. 

\noindent \textbf{Image-to-image translation.} Image translation has gained considerable attention generative adversarial networks (GANs)~\cite{CycleGAN2017,ugatit2020}. Though the GAN method is originally derived for the unsupervised setting, the supervised learning method for GANs that conditions the latent space on the label space was suggested in~\cite{zhu2017unpaired}.
\begin{figure}[!t]
\centering
\includegraphics[width=1\linewidth]{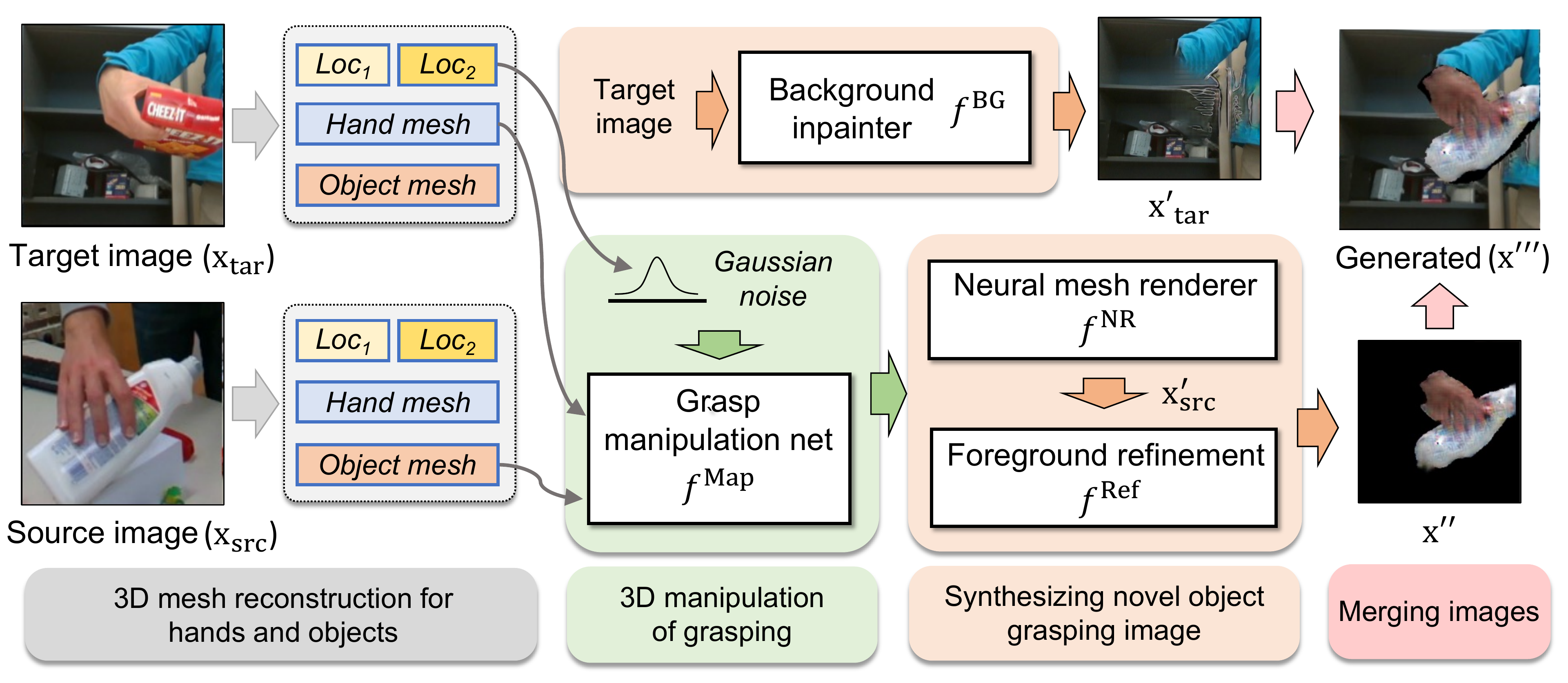}
  \caption{A schematic diagram of the proposed hand-object interaction reenactment framework: First, hand and object meshes are inferred. Then, the source object and target hand are used to infer the tight hand grasps via the grasp manipulation network $f^{\text{Map}}$. The 2D image $\mbx'_{\text{src}}$ is generated via projecting 3D meshes using neural mesh renderer $f^{\text{NR}}$ and it is refined towards $\mbx''$ by the foreground refinement network $f^\text{Ref}$ to have the realistic quality. At the same time, the original target image $\mbx_{\text{tar}}$ is used to generate the background image $\mbx'_{\text{tar}}$ by inpainting removed foreground regions (i.e., hands, objects). Finally, $\mbx'''$ is generated by merging $\mbx''$ and $\mbx'_{\text{tar}}$.}
  \vspace{-0.3cm}
\label{fig:diagram}
\end{figure}



\section{Pipeline}

 Our aim here is to achieve the \textit{hand-object grasp reenactment} task that exchanges objects of source and target images, learn to find proper hand grasps for the exchanged object, and generates quality 2D images. Towards that direction, we defined 4 steps in our pipeline as in Fig.~\ref{fig:diagram}, which are detailed in Sec.~\ref{sec:method1} through Sec.~\ref{merge}. 

\subsection{3D mesh reconstruction for hands and objects}
\label{sec:method1}

We first proposed to estimate 3D meshes of hands and objects (i.e. $\mbm=\{\mbm_{\text{H}},\mbm_{\text{O}}\}$), location parameters for hand meshes (i.e., $\mbc_1$, $\mbc_2$) given an RGB image $\mbx$ (cf. we used two types of location parameters, $\mbc=[\mbc_1$, $\mbc_2]$, which are used to transform the mesh within the original MANO~\cite{MANO:SIGGRAPHASIA:2017} space and transform the mesh from MANO space to the image space, respectively). 
Additionally, to infer textures of hand and object meshes, we involved the texture inference network $f^{\text{Tex}}$, which infers RGB texture vector $\mbt$ corresponding to faces of the 3D meshes $\mbm=\{\mbm_{\text{H}}, \mbm_{\text{O}}\}$.

\begin{figure}[t]
\centering
\captionsetup[subfigure]{labelformat=empty}
\subfloat[(a)]{\includegraphics[width=0.16\linewidth]{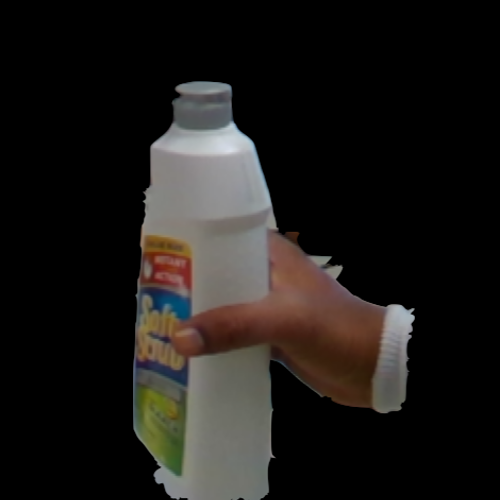}}
\subfloat[(b)]{\includegraphics[width=0.16\linewidth]{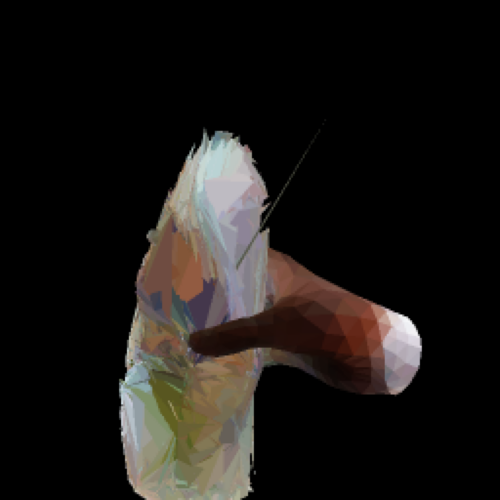}}
\subfloat[(c)]{\includegraphics[width=0.16\linewidth]{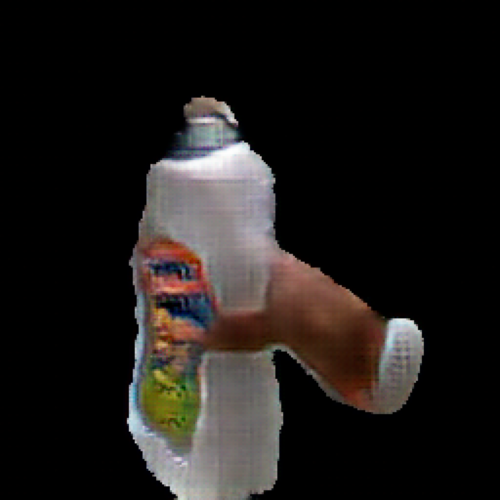}}
\hspace{0.3em}
\subfloat[(d)]{\includegraphics[width=0.16\linewidth]{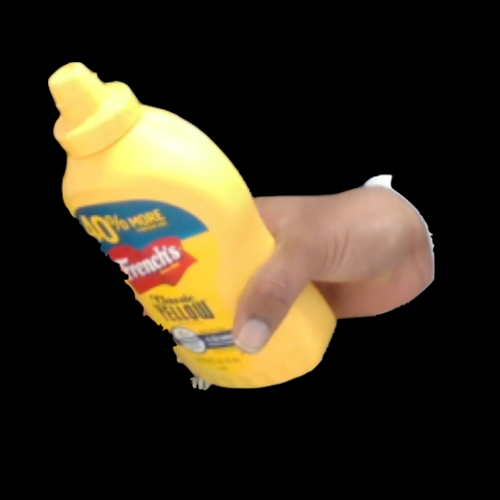}}
\subfloat[(e)]{\includegraphics[width=0.16\linewidth]{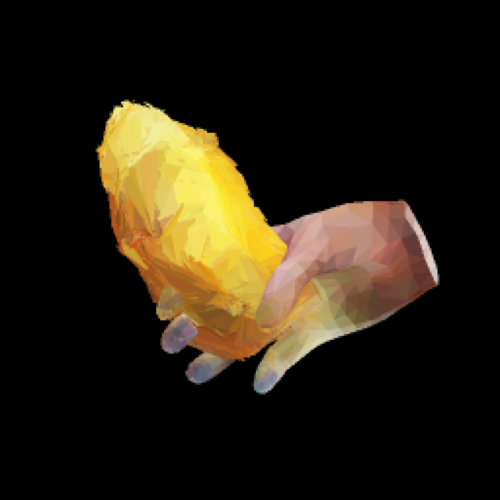}}
\subfloat[(f)]{\includegraphics[width=0.16\linewidth]{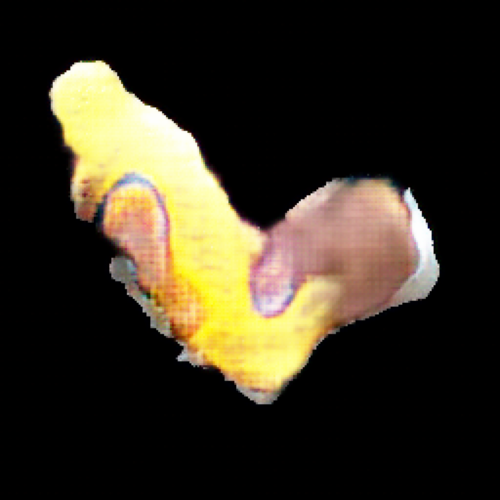}}
\caption{Ablation study for with and without the use of the foreground refinement network $f^{\text{Ref}}$. (a, d) Ground-truth image, (b, e) Generated image before $f^{\text{Ref}}$, (c, f) Refined image after $f^{\text{Ref}}$.}
\label{fig:ab1}
\end{figure}

\begin{figure}
\centering
\captionsetup[subfigure]{labelformat=empty}
\subfloat[(a)]{\includegraphics[width=0.46\linewidth]{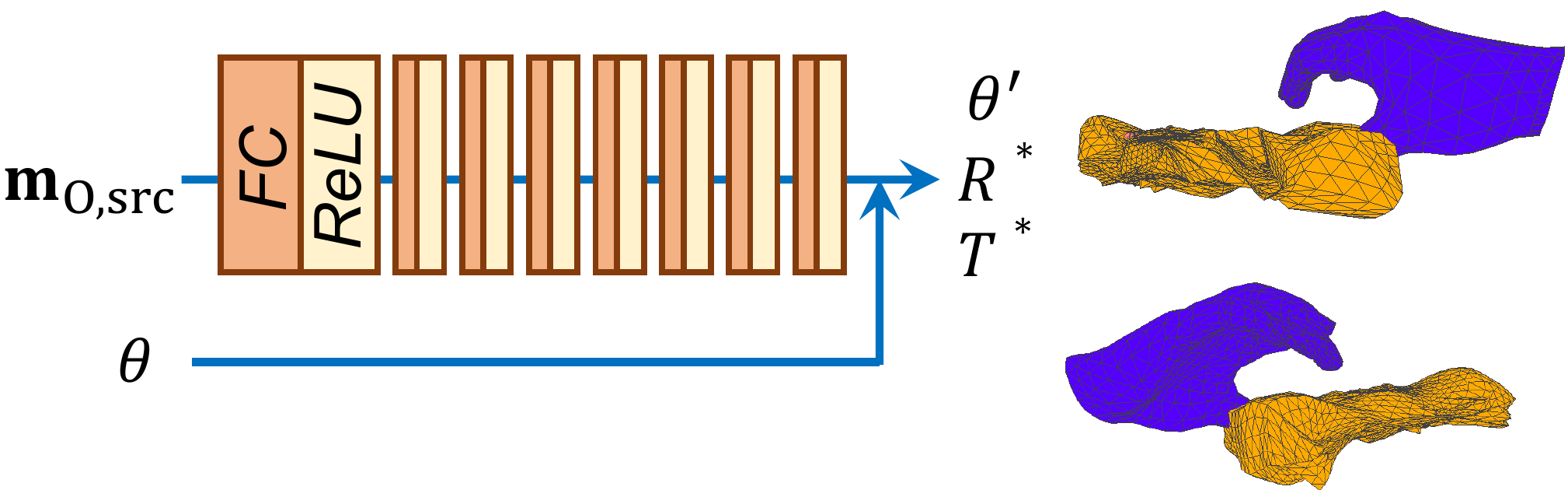}} \hspace{0.3cm}
\subfloat[(b)]{\includegraphics[width=0.46\linewidth]{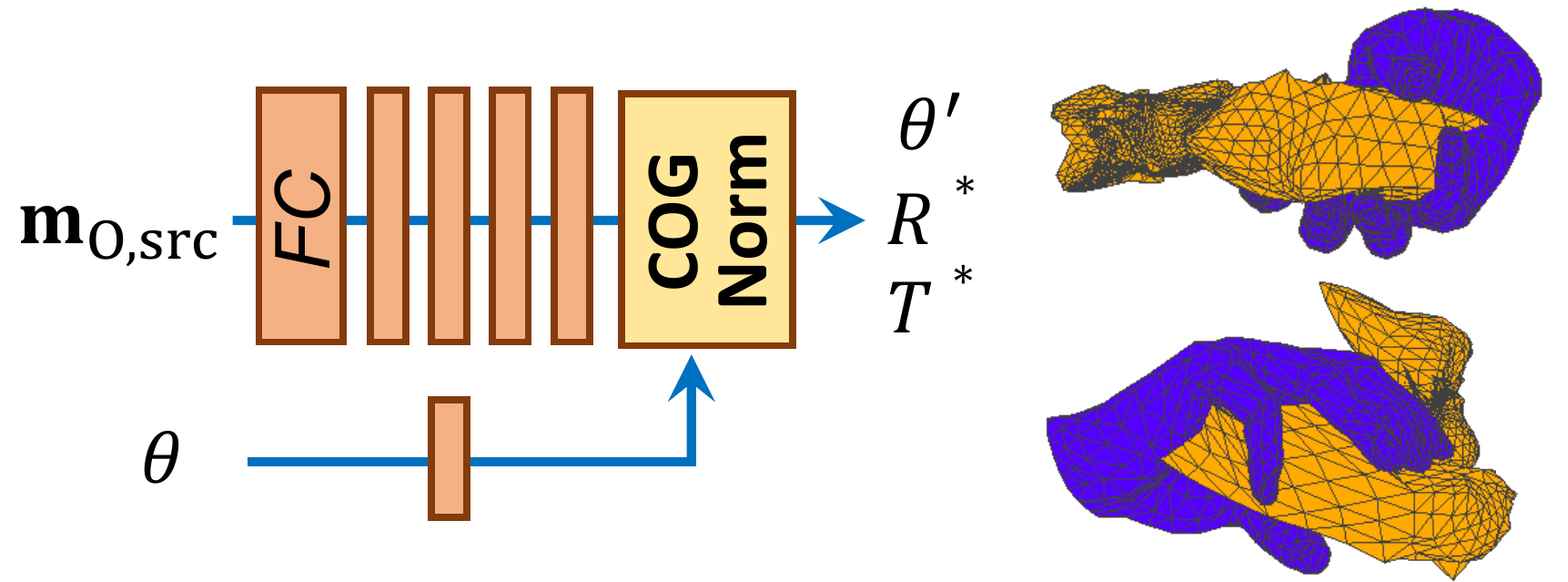}}
\caption{Ablation study on the use of COG-Norm layer. (a) the parameter size of the grasp manipulation network $f^{\text{Map}}$ having only MLP layers is 0.97M; (b) the parameter size of the grasp manipulation network $f^{\text{Map}}$ with COG-Norm layer is 0.49M. Also, we can observe that in (b), reconstructed hand-object meshes are more tightly aligned compared to the one in (a).}
\label{fig:ab2}
\end{figure}

\noindent \textbf{Hand mesh reconstruction network $f^{\text{Hand}}$.} The hand mesh reconstruction network $f^{\text{Hand}}: \mbx \rightarrow [\mbm_{\text{H}}(\theta, \beta), \mbc]$ is defined to input the 2D image $\mbx$ and output the location parameters $\mbc=[\mbc_1, \mbc_2]$ and 3D hand mesh $\mbm_{\text{H}}(\theta, \beta)\in\mathbb{R}^{778\times 1,538}$. The 3D hand mesh $\mbm_{\text{H}}(\theta, \beta)$ is parameterized by $\theta$ and $\beta$, which denotes pose and shape parameters of the MANO model, respectively~\cite{MANO:SIGGRAPHASIA:2017}. Both location parameters $\mbc_1$ and $\mbc_2$ have seven scalar values: $1$ for scale, $2$ for x and y translation, and $4$ for quaternion, respectively. We employed the architecture of the convolutional pose machine (CPM)~\cite{wei2016cpm} combined with the MANO layer~\cite{MANO:SIGGRAPHASIA:2017} to first estimate 2D heatmaps of the hands and then estimate the 3D hand meshes upon it. 



\noindent \textbf{Object mesh reconstruction network $f^{\text{Object}}$.} Our 3D object mesh reconstruction network $f^{\text{Object}}:\mbx\rightarrow \mbm_{\text{O}}\in\mathbb{R}^{2,466\times 4,928}$ used the same architecture as Pixel2Mesh~\cite{wang2018pixel2mesh}.



\noindent \textbf{Texture inference network $f^{\text{Tex}}$.} Our texture inference network $f^{\text{Tex}}: \mbx \rightarrow \mbt\in \mathbb{R}^{N_f\times T\times T\times T\times 3}$ uses the ResNet-50 architecture to infer the RGB texture vector $\mbt$ that corresponds to each face of 3D meshes $\mbm$. 
$T$ is the minimum dimension for  $\mbt$ and it is set to $2$.

\subsection{3D manipulation of novel hand grasping}
\label{sec:method2}
To adjust hand postures of the target image $\mbx_{\text{tar}}$ to tightly grasp the translated 3D object meshes $\mbm_{\text{O},\text{src}}$ of the source image $\mbx_{\text{src}}$, we proposed a grasp manipulation network $f^{\text{Map}}$ that manipulates the pose parameter $\theta$ of the initial hand mesh $\mbm_{\text{H},\text{tar}}$. We propose the \textbf{CO}ntact-aware \textbf{G}rasp \textbf{Norm}alization (\textbf{COG-Norm}) layer to effectively output a new hand pose parameter $\theta'$ for $\mbm_{\text{H},\text{tar}}$ by mixing vertice information of predicted hands and objects.

\noindent \textbf{Grasp manipulation network $f^{\text{Map}}$.} In the COG-Norm layer, the pose parameter is normalized and modulated with the features of $\mbm_{\text{O},\text{src}}$ obtained by the function $\gamma$ and $\alpha$, which are composed of fully-connected layers.
The activation formulation is calculated as below ($i \in [1, 45]$):
\begin{eqnarray}
\text{COG-Norm}(\theta_i, \mbm_{\text{O},\text{src}}) &=& \gamma(\mbm_{\text{O},\text{src}})\frac{\theta_i - \mu(\theta)}{\sigma(\theta)} \nonumber\\&+& \alpha(\mbm_{\text{O},\text{src}})
\end{eqnarray}
where $\theta_i$, $\mu$ and $\sigma$ are the $i$-th element of the $\theta$, the mean and standard deviation of MANO pose parameters.


\subsection{Synthesizing novel object grasping image}
\label{sec:method3}

\noindent \textbf{Neural 3D mesh renderer $f^{\text{NR}}$.} Neural 3D mesh renderer $f^{\text{NR}}: [\mbm, \mbt, \mbc_2]\rightarrow \mbx'$ is used to render the reconstructed 3D meshes $\mbm$ and its texture vector $\mbt$ to the 2D image $\mbx'$. 
We employ the implementation of~\cite{kato2018nmr} for this.

\noindent \textbf{Foreground refinement network $f^{\text{Ref}}$.} We proposed to employ the foreground refinement network   $f^{\text{Ref}}:\mbx'\rightarrow \mbx''$ to refine the initial 2D images $\mbx'$ towards more realistic 2D images $\mbx''$. Its architecture is same as the generation network of~\cite{wang2018pix2pixHD}.

\noindent \textbf{Background in-painter $f^{\text{BG}}$.} Background in-painter is used to reuse the background of the target image $\mbx_{\text{tar}}$. We propose to remove the foreground objects in the target image $\mbx_{\text{tar}}$ using the inpainting algorithm: 1) we re-projected the estimated hand and object 3D meshes $\mbm$ to generate the 2D foreground mask $\mbs_{\text{tar}}$ having binary values ($1$ for foreground objects, $0$ for backgrounds), 2) then we applied the in-painting network of~\cite{yi2020contextual} pre-trained on Places2~\cite{zhou2017places} dataset onto $(1-\mbs_{\text{tar}})\odot \mbx_{\text{tar}}$ to obtain the in-painted image $\mbx'_{\text{tar}}$.

\subsection{Merging images.}\label{merge} The final image $\mbx'''$ is generated by merging the refined image $\mbx''$ and background-inpainted image $\mbx'_{\text{tar}}$ using the foreground mask $\mbs$ rendered from hand and object meshes $[\mbm_{\text{H,tar}}(\theta'), \mbm_\text{O,src}]$:
\begin{eqnarray}
\mbx''' = \mbx''\odot\mbs + \mbx'_{\text{tar}}\odot(1-\mbs)
\end{eqnarray}
where $\odot$ denotes the element-wise multiplication operation.

\noindent \textbf{Aligning wrist positions.} When aligning arms in the target image $\mbx_{\text{tar}}$ with the transferred hand postures, we choose the 1 wrist point of the hand mesh $\mbm_{\text{H}}$ that is estimated from $\mbx_{\text{tar}}$. Then, we translate the absolute position of the transferred hand and object meshes by making its wrist point coincide with the kept wrist point. 


\section{Training method}


Our training method is composed of two steps: 1) a step for training $f^{\text{Hand}}$, $f^{\text{Object}}$, $f^{\text{Tex}}$, $f^{\text{Ref}}$ and 2) a step for training $f^{\text{Map}}$. At the first step, we use the loss function $\mathcal{L}_{\text{HO}}$ to train four networks: $f^{\text{Hand}}$, $f^{\text{Object}}$, $f^{\text{Tex}}$, $f^{\text{Ref}}$: 
\begin{eqnarray}
&&\mathcal{L}_{\text{HO}}(f^{\text{Hand}}, f^{\text{Object}}, f^{\text{Tex}}, f^{\text{Ref}}) = \mathcal{L}_{\text{Hand}}(f^{\text{Hand}})\nonumber\\&&+ \lambda_{\text{ho},1} \mathcal{L}_{\text{Object}}(f^{\text{Object}})\nonumber+ \lambda_{\text{ho},2} \mathcal{L}_{\text{Tex}}(f^{\text{Tex}})+ \lambda_{\text{ho},3} \mathcal{L}_{\text{Ref}}(f^{\text{Ref}})
\end{eqnarray}
where $\lambda_{\text{ho},1}=1$, $\lambda_{\text{ho},2}=0.01$, and $\lambda_{\text{ho},3}=0.1$ are used to balance each term. Then, we train the $f^{\text{Map}}$ using the loss $\mathcal{L}_{\text{Map}}$ in Eq.~\ref{eq:lmap}. The first step is executed for $30$ epochs and the second step is processed for $15$ epochs. We trained the entire network using the Adam optimizer with $\beta=(0.5, 0.999)$ and a learning rate of $10^{-4}$ for hand and object reconstruction networks: $f^{\text{Hand}}$, $f^{\text{Object}}$ while $10^{-5}$ for all others. Each loss term is detailed in the remainder of this section: 


\begin{table*}[t!]
\centering
\caption{Quantitative results comparing ours to three state-of-the-art image translation methods (i.e., CycleGAN~\cite{CycleGAN2017}, U-GAT-IT~\cite{ugatit2020}, and ReenactGAN~\cite{wu2018reenactgan}) on Honnotate~\cite{hampali2020honnotate} and DexYCB~\cite{dexycb:cvpr2021} datasets. The evaluation is performed using three measures: FID~\cite{fid_nips}, LPIPS~\cite{lpips_cvpr} based on AlexNet, and LPIPS based on VGG. For all measures, lower is better and best results are bold-faced. 
}
{\renewcommand{\arraystretch}{1.3}
\resizebox{\textwidth}{!}{
\begin{tabular*}{1.2\textwidth}{cccc|ccc}
\multicolumn{1}{c|}{DB} && Honnotate~\cite{hampali2020honnotate} &&& DexYCB~\cite{dexycb:cvpr2021} & \\ \Xhline{3\arrayrulewidth} \multicolumn{1}{c|}{\diagbox[innerwidth=1.5cm,innerrightsep=0pt, innerleftsep=0pt]{Model}{Stream}}& \begin{tabular}[c]{@{}c@{}}$\mathsf{A} \rightarrow \mathsf{B}$\\$\mathsf{B} \rightarrow \mathsf{A}$\end{tabular}& \begin{tabular}[c]{@{}c@{}}$\mathsf{A} \rightarrow \mathsf{C}$\\$\mathsf{C} \rightarrow \mathsf{A}$\end{tabular} & \begin{tabular}[c]{@{}c@{}}$\mathsf{B} \rightarrow \mathsf{C}$\\$\mathsf{C} \rightarrow \mathsf{B}$\end{tabular} &
\begin{tabular}[c]{@{}c@{}}$\mathsf{A} \rightarrow \mathsf{B}$\\$\mathsf{B} \rightarrow \mathsf{A}$\end{tabular}& \begin{tabular}[c]{@{}c@{}}$\mathsf{A} \rightarrow \mathsf{C}$\\$\mathsf{C} \rightarrow \mathsf{A}$\end{tabular} & \begin{tabular}[c]{@{}c@{}}$\mathsf{B} \rightarrow \mathsf{C}$\\$\mathsf{C} \rightarrow \mathsf{B}$\end{tabular} \\ \Xhline{3\arrayrulewidth}
\multicolumn{1}{c|}{\begin{tabular}[c]{@{}c@{}}CycleGAN\end{tabular}} &\begin{tabular}[c]{@{}c@{}}914.4 / 0.478 / 0.583 \\ 887.8 / 0.575 / 0.632\end{tabular}  &\begin{tabular}[c]{@{}c@{}}843.2 / 0.546 / 0.611 \\ \textbf{839.2} / 0.584 / 0.637\end{tabular}& \begin{tabular}[c]{@{}c@{}}915.6 / 0.583 / 0.661 \\ 984.9 / 0.590 / 0.608\end{tabular} & \begin{tabular}[c]{@{}c@{}}509.8 / 0.227 / 0.361 \\ 758.6 / 0.255 / 0.377\end{tabular} & \begin{tabular}[c]{@{}c@{}}497.0 / 0.244 / 0.353 \\ \textbf{502.4} / 0.179 / 0.290\end{tabular}  & \begin{tabular}[c]{@{}c@{}}627.0 / 0.180 / 0.277 \\ 612.3 / 0.208 / 0.343\end{tabular}\\
\multicolumn{1}{c|}{\begin{tabular}[c]{@{}c@{}}U-GAT-IT\end{tabular}} & \begin{tabular}[c]{@{}c@{}}\textbf{734.2} / 0.443 / 0.533 \\ 781.8 / 0.526 / 0.631\end{tabular} & \begin{tabular}[c]{@{}c@{}}788.1 / 0.624 / 0.689 \\ 872.0 / 0.630 / 0.693\end{tabular}  & \begin{tabular}[c]{@{}c@{}}864.0 / 0.612 / 0.696 \\ 1011.3 / 0.640 / 0.688\end{tabular} &  \begin{tabular}[c]{@{}c@{}}502.0 / 0.269 / 0.411 \\ 561.5 / 0.252 / 0.387\end{tabular} & \begin{tabular}[c]{@{}c@{}}496.3 / 0.285 / 0.430 \\ 477.0 / 0.261 / 0.400\end{tabular}  & \begin{tabular}[c]{@{}c@{}}\textbf{490.0} / 0.211 / 0.347 \\ \textbf{519.0} / 0.251 / 0.406\end{tabular}\\
\multicolumn{1}{c|}{\begin{tabular}[c]{@{}c@{}}ReenactGAN\end{tabular}} & \begin{tabular}[c]{@{}c@{}}1845.7 / 0.482 / 0.466 \\ 1530.8 / 0.370 / 0.415\end{tabular} & \begin{tabular}[c]{@{}c@{}}1491.1 / 0.440 / 0.381 \\ 1470.6 / 0.372 / 0.414\end{tabular}  & \begin{tabular}[c]{@{}c@{}}1707.1 / 0.372 / 0.376 \\ 1732.6 / 0.364 / 0.422\end{tabular} &  \begin{tabular}[c]{@{}c@{}}1102.9 / 0.484 / 0.473 \\ 1806.3 / 0.500 / 0.454\end{tabular} & \begin{tabular}[c]{@{}c@{}}1268.3 / 0.519 / 0.496 \\ 1917.0 / 0.532 / 0.454\end{tabular}  & \begin{tabular}[c]{@{}c@{}}1313.0 / 0.521 / 0.472 \\ 1447.8 / 0.512 / 0.467\end{tabular}\\


\multicolumn{1}{c|}{\begin{tabular}[c]{@{}c@{}}Ours\\(w/o $f^\text{Ref}$)\end{tabular}} & \begin{tabular}[c]{@{}c@{}}862.9 / 0.713 / 0.541 \\ 983.1 / 0.768 / 0.617\end{tabular}  & \begin{tabular}[c]{@{}c@{}}828.9 / 0.674 / 0.528 \\ 917.6 / 0.634 / 0.565\end{tabular} & \begin{tabular}[c]{@{}c@{}}997.7 / 0.771 / 0.618 \\ 939.3 / 0.637 / 0.573\end{tabular} & \begin{tabular}[c]{@{}c@{}}443.0 / 0.227 / 0.210 \\ 585.9 / 0.154 / 0.214\end{tabular} & \begin{tabular}[c]{@{}c@{}}483.7 / 0.153 / 0.212 \\ 545.3 / 0.249 / 0.304\end{tabular}  & \begin{tabular}[c]{@{}c@{}}558.2 / 0.151 / 0.207 \\ 536.5 / 0.258 / 0.316\end{tabular}
\\

\multicolumn{1}{c|}{Ours} & \begin{tabular}[c]{@{}c@{}}777.6 / \textbf{0.417} / \textbf{0.397} \\ \textbf{585.6} / \textbf{0.168} / \textbf{0.167}\end{tabular}  & \begin{tabular}[c]{@{}c@{}}\textbf{752.7} / \textbf{0.383} / \textbf{0.368} \\ 851.9 / \textbf{0.242} / \textbf{0.248}\end{tabular} & \begin{tabular}[c]{@{}c@{}}\textbf{636.9} / \textbf{0.184} / \textbf{0.180} \\ \textbf{891.7} / \textbf{0.285} / \textbf{0.286}\end{tabular}&\begin{tabular}[c]{@{}c@{}}\textbf{404.1} / \textbf{0.132} / \textbf{0.197} \\ \textbf{525.1} / \textbf{0.138} / \textbf{0.191}\end{tabular} & \begin{tabular}[c]{@{}c@{}}\textbf{399.0} / \textbf{0.130} / \textbf{0.197} \\ 517.1 / \textbf{0.176} / \textbf{0.218}\end{tabular}  & \begin{tabular}[c]{@{}c@{}}522.2 / \textbf{0.144} / \textbf{0.192} \\ 542.3 / \textbf{0.177} / \textbf{0.223}\end{tabular}
\end{tabular*}}}
\label{tab:result1}
\end{table*}

\noindent \textbf{Hand mesh estimation loss $\mathcal{L}_{\text{Hand}}$.} This loss is used to train the hand mesh reconstruction network $f^{\text{Hand}}$ by 1) closing hand meshes $\mbm_{\text{H}}$ to its ground-truths $\mbm_{\text{H,GT}}$ and, 2) closing 3D skeletons $J(\mbm_{\text{H}})$ and 2D skeletons $\psi_{\mbc_{\text{2}}}(J(\mbm_{\text{H}}))$ regressed from hand meshes to their ground-truths $\mbj_{\text{GT}}$ and $\mbj_{\text{2DGT}}$, respectively:
\begin{eqnarray}
&&\mathcal{L}_{\text{Hand}}(f^{\text{Hand}}) = \|\psi_{\mbc_{\text{1}}}(\mbm_{\text{H}}) - \psi_{\mbc_{\text{1}}}(\mbm_{\text{H,GT}})\|_1 \nonumber\\&&+ \|\psi_{\mbc_{\text{1}}}(J(\mbm_{\text{H}})) - \mbj_{\text{GT}}\|_2^2 + ||\psi_{\mbc_{\text{2}}}(J(\mbm_{\text{H}})) - \mbj_{\text{2DGT}}||_2^2
\end{eqnarray}
where the projection function $\psi_{\mbc}$ applies scale, translation, and rotation transformations to 3D meshes using $\mbc$. 
The network $J(\cdot)$ provided by~\cite{MANO:SIGGRAPHASIA:2017} geometrically regresses 3D skeletons $\mbj$ from mesh $\mbm_{\text{H}}$.

\noindent \textbf{Object mesh estimation loss $\mathcal{L}_{\text{Object}}$.} This loss is defined as follows:
\begin{eqnarray}
\mathcal{L}_{\text{Object}}(f^{\text{Object}})= L_{\text{chamfer}}(f^{\text{Object}})+ L_{\text{edge}}(f^{\text{Object}})+L_{\text{lap}}(f^{\text{Object}})
\end{eqnarray}
where chamfer loss $L_{\text{chamfer}} = \sum_p \min_q ||p-q||_2^2 + \sum_q \min_p ||p-q||_2^2$, edge loss $L_{\text{edge}} = \sum_{p} \sum_{k \in N(p)}||p-k||_2^2$, and Laplacian loss $L_{\text{lap}} = \sum_{p}||\delta_p' - \delta_p||_2^2$ are employed from~\cite{wang2018pixel2mesh} to train the 3D object mesh reconstruction network.


\noindent \textbf{Texture inference loss $\mathcal{L_{\text{Tex}}}$.} This loss is used to train the texture inference network $f^{\text{Tex}}$ by making the rendered RGB images from hand-object meshes with the inferred textures $f^{\text{Tex}}(\mbx)$ same as the segmented foreground regions of the original image $\mbx$:
 \begin{eqnarray}
 \mathcal{L}_{\text{Tex}}(f^{\text{Tex}}) = \|{f^{\text{NR}}}([\mbm_{\text{H}}, \mbm_{\text{O}}], f^{\text{Tex}}(\mbx), \mbc_{\text{2}}) - \mbx\odot \mbs'\|_2^2
 \end{eqnarray}
where $\mbs'$ is a foreground mask rendered from meshes $\mbm=[\mbm_{\text{H}}, \mbm_{\text{O}}]$.

\noindent \textbf{Refinement loss $\mathcal{L}_{\text{Ref}}$.} This loss is used to train the foreground refinement network $f^{\text{Ref}}$ to generate more realistic images $\mbx''=f^{\text{Ref}}(\mbx'))$ from $\mbx'$. 
\begin{eqnarray}
\mathcal{L}_{\text{Ref}}(f^{\text{Ref}})=L_{\text{GAN}}(f^{\text{Ref}})+L_{\text{P}}(f^{\text{Ref}})
\end{eqnarray}
where 
\begin{eqnarray}
L_{\text{GAN}}(f^{\text{Ref}}) &=& \mathbb{E}_{(\mbx',\mbx)}[\log D^{\text{Ref}}(\mbx',\mbx)]  \nonumber\\ &+& \mathbb{E}_{\mbx'}[\log(1-D^{\text{Ref}}(\mbx',f^{\text{Ref}}(\mbx'))],\\
L_{\text{P}}(f^{\text{Ref}}) &=&
\sum_{i=1}^{N} \frac{1}{M_i}[\|F^{(i)}(\mbx) - F^{(i)}(f^{\text{Ref}}(\mbx'))\|_1]
\end{eqnarray}
where $D^\text{Ref}$ is the GAN-type discriminator network~\cite{CycleGAN2017} and $F^{(i)}$ denotes the $i$-th layer of the VGG-19 network having $M_i$ elements which is involved to capture the perceptual characteristics~\cite{lpips_cvpr}.

\noindent \textbf{Grasp manipulation loss $\mathcal{L}_{\text{Map}}$.} This loss is used to train the grasp manipulation network $f^{\text{Map}}$ that learns to control the hand postures to grasp the objects tightly:

\begin{eqnarray}
\mathcal{L}_{\text{Map}}(f^{\text{Map}})&=&L_{\text{contact}}(f^{\text{Map}}) + L_{\text{cent}}(f^{\text{Map}}) + L_{\text{adv}}(f^{\text{Map}})
\label{eq:lmap}
\end{eqnarray}
where
\begin{eqnarray}
L_{\text{contact}}(f^{\text{Map}}) &=& \frac{1}{|\mbv_{\text{contact}}|} \sum_{v \in \mbv_{\text{contact}}} \min_k \|v, \mbm_{\text{O}}^{k}\|_2\\
L_{\text{cent}}(f^{\text{Map}})&=&\frac{1}{2} \| \mathbb{C}(\mbm_{\text{O}}) - \mathbb{C}(\mbm_{\text{H}})\|_2^2\\
L_{\text{adv}}(f^{\text{Map}}) &=& -\mathbb{E}_{\theta}[D(\theta)]  + \mathbb{E}_{\theta_{\text{GT}}}[D(\theta_{\text{GT}})] \nonumber\\&+& \lambda \mathbb{E}_{\theta}[(\|\nabla_{\theta}D(\theta)\|_2 - 1)^2]
\end{eqnarray} 
where contact loss $L_{\text{contact}}$ is used to minimize the distance between 3D object vertices $\mbm_{\text{O}}^{k}$ and pre-defined vertices of hands $\mbv_{\text{contact}}$ that are frequently involved for the grasping, centroid loss $L_{\text{cent}}$ is used to minimize the center position of hand and objects where $\mathbb{C}(\cdot)$ denotes the centroid of 3D meshes and the adversarial loss $L_{\text{adv}}$ is involved to constitute the adversarial~\cite{CycleGAN2017} and Wassertein~\cite{wassertain} losses via the discriminator $D$, respectively. Via $L_\text{contact}$ and $L_\text{centroid}$, $f^{\text{Map}}$ is learned to obtain the tight 3D grasp between objects and hand meshes. Also, $L_\text{adv}$ enforces $f^{\text{Map}}$ to obtain the hand grasps within the realistic grasping distribution.

\begin{figure*}[t!]
\centering
\includegraphics[width=1\linewidth]{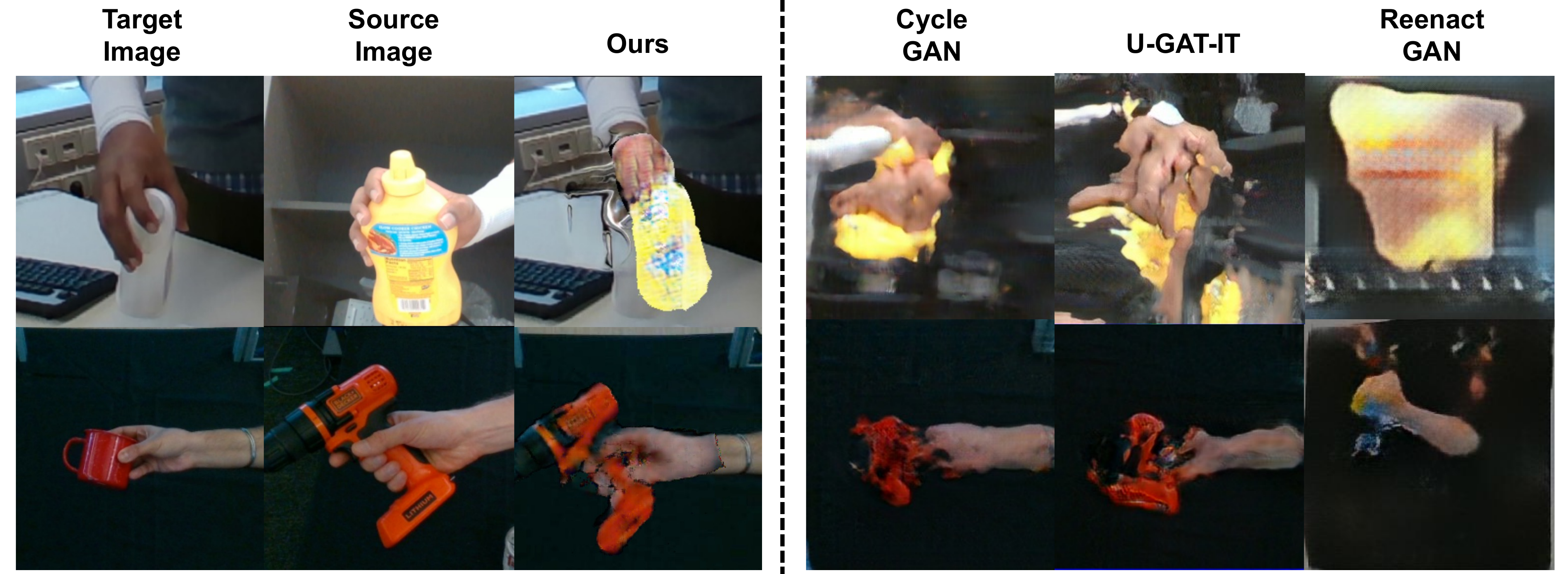}
   \caption{Examples from Ours and other baselines (CycleGAN, U-GAT-IT and ReenactGAN) on Honnotate~\cite{hampali2020honnotate}~(first rows) and DexYCB~\cite{dexycb:cvpr2021}~(last rows). Ours faithfully generates 2D images; while others show incomplete results.}
\label{fig:fig3}
\end{figure*}

\section{Experiment}

\noindent \textbf{Dataset.} We conducted our experiments on two types of datasets: Honnotate~\cite{hampali2020honnotate} and DexYCB~\cite{dexycb:cvpr2021}. 
For Honnotate dataset, we used the entire dataset, while for DexYCB dataset, we used the default split: sub-sampling ten out of twenty subjects, one out of eight views excluding non-grasping frames when constituting the dataset. 


\noindent \textbf{Results.} We involved two conventional image translation methods (i.e., CycleGAN~\cite{CycleGAN2017} and U-GAT-IT~\cite{ugatit2020}) with one face reenactment method (i.e., ReenactGAN~\cite{wu2018reenactgan}) to compare with our method. We found that ours obtained the superior performance in three measures (i.e., FID~\cite{fid_nips}, LPIPS~\cite{lpips_cvpr} based on AlexNet and LPIPS based on VGG) compared to three algorithms as shown in Table~\ref{tab:result1}. For the binary domain transfer, we picked three objects as domains: (A) potted meat can, (B) bleach cleanser, and (C) mustard bottle and trained the model for the bi-directional transfer (i.e., $A\rightarrow B$, $B\rightarrow A$, $A\rightarrow C$, $C\rightarrow A$, $B\rightarrow C$, and $C\rightarrow B$). 


Fig.~\ref{fig:fig3} shows the qualitative examples for two datasets. We visualize the results from 2D image translation methods (ie. CycleGAN~\cite{CycleGAN2017}, U-GAT-IT~\cite{ugatit2020}), denoting that the 2D image manipulation models are not able to understand the foreground objects. We also compared ours with reenactment algorithm (ie. ReenactGAN~\cite{wu2018reenactgan}) as well; however it fails to deliver the reenactment of hand-object grasping, even though it exploits 2D poses of the foreground objects. On the contrary, ours faithfully maintain the background of target images while changing only foreground pixels. 


\noindent \textbf{Ablation study.} We additionally conducted ablation studies for our foreground refinement network $f^{\text{Ref}}$ and COG-Norm layer proposed in the grasp manipulation network $f^{\text{Map}}$. In Fig.~\ref{fig:ab1} and Table.~\ref{tab:result1}, we showed the results with and without the foreground refinement network $f^{\text{Ref}}$: The result is quantitatively and qualitatively better when $f^{\text{Ref}}$ is involved. In Fig.~\ref{fig:ab2}, we attached qualitative results comparing the grasp manipulation network $f^{\text{Map}}$ constructed using only FC layers versus the one using the proposed COG-Norm layer. From this, we could find that the use of the COG-Norm layer makes a huge difference: it makes the entire grasp manipulation network $f^{\text{Map}}$ work better and faster than before. 

\section{Conclusion}
We have introduced the novel scenario of hand-object grasp reenactment and the HOReeNet, a fully differentiable pipeline that is able to deal with the scenario. 
It performs better than conventional image translation and face reenactment algorithms on the novel task. From ablative experiments, we also observed that proposed sub-modules operate in the meaningful way. We expect the proposed method and scenario could be the stepping stone for future research that offers 3D information in 2D-based generative networks. 

\vfill\pagebreak


\end{document}